\PassOptionsToPackage{table}{xcolor}
\documentclass[10pt,twocolumn,letterpaper]{article}
\usepackage{cvpr}              
\usepackage{graphicx}
\usepackage{amsmath}
\usepackage{amssymb}
\usepackage{booktabs}
\usepackage{makecell}  
\usepackage{multirow}
\usepackage{textcomp}
\usepackage{algorithm}
\usepackage{algpseudocode}  
\usepackage{booktabs}
\usepackage{multirow}
\usepackage{amsmath, amsfonts, amssymb}
\usepackage{tabularx}   
\usepackage{array}      
\usepackage{amsthm}   

\newtheorem{definition}{Definition}

\newtheorem{theorem}{Theorem}
\newtheorem{remark}{Remark}
\newtheorem{assumption}{Assumption}

\usepackage{comment}
\usepackage[dvipsnames]{xcolor}

\definecolor{cvprblue}{rgb}{0.21,0.49,0.74}
\usepackage[pagebackref,breaklinks,colorlinks,citecolor=cvprblue]{hyperref}

\def\paperID{7631} 
\def\confName{CVPR}
\def\confYear{2024}

\title{EVTP-IVS: Effective Visual Token Pruning For Unifying Instruction Visual Segmentation In Multi-Modal Large Language Models}

\author{Wenhui Zhu$^{13\dag*}$ \quad Xiwen Chen$^{2*}$ \quad Zhipeng Wang$^{3*\ddag}$ \quad Shao Tang$^{3}$  \quad Sayan Ghosh$^{3}$ \quad Xuanzhao Dong$^{1}$ \\ \quad Rajat Koner$^{4}$ \quad Yalin Wang$^{1}$ \\
\\
$^{1}$ Arizona State University, AZ, USA \\  
$^{2}$ Clemson University, SC, USA \\ 
$^{3}$ LinkedIn Corporation, CA, USA \\
$^{4}$ Ludwig Maximilian University of Munich, Munich, Germany \\
}

\begin{document}
\maketitle
\def\thefootnote{*}\footnotetext{ Contributed equally to this paper.}
\def\thefootnote{\dag}\footnotetext{This work was conducted during the internship at LinkedIn.}
\def\thefootnote{\ddag}\footnotetext{Corresponding author at zhipwang@linkedin.com }

\begin{abstract}
Instructed Visual Segmentation (IVS) tasks require segmenting objects in images or videos based on natural language instructions. While recent multimodal large language models (MLLMs) have achieved strong performance on IVS, their inference cost remains a major bottleneck, particularly in video. We empirically analyze visual token sampling in MLLMs and observe a strong correlation between subset token coverage and segmentation performance. This motivates our design of a simple and effective token pruning method that selects a compact yet spatially representative subset of tokens to accelerate inference. In this paper, we introduce a novel visual token pruning method for IVS, called EVTP-IV, which builds upon the $k$-center by integrating spatial information to ensure better coverage. We further provide an information-theoretic analysis to support our design. Experiments on standard IVS benchmarks show that our method achieves up to 5× speed-up on video tasks and 3.5× on image tasks, while maintaining comparable accuracy using only 20\% of the tokens. Our method also consistently outperforms state-of-the-art pruning baselines under varying pruning ratios.
\end{abstract}

\begin{figure}[t]
  \centering
  \includegraphics[width=0.5\textwidth]{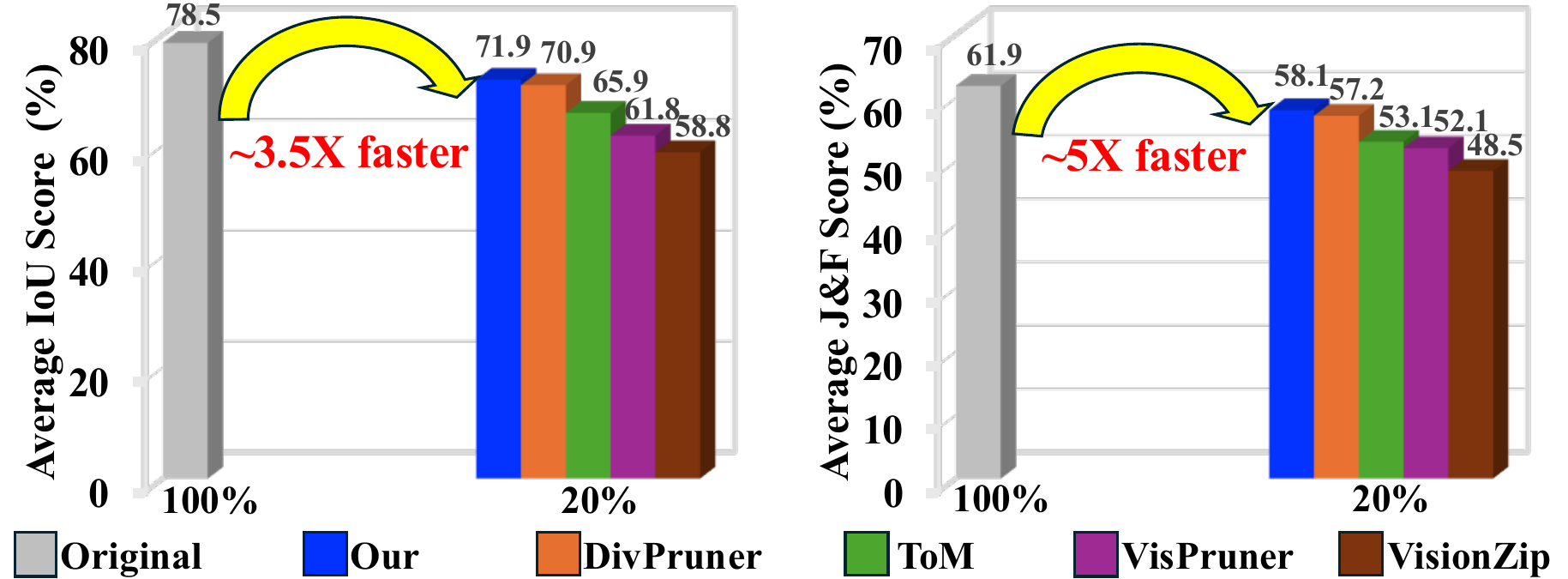}
  \caption{\textbf{Comparison of segmentation accuracy and inference speed on IVS benchmarks. (Left: all Image, Right: all Video)} Our method achieves significantly faster inference (3.5× for images, 5× for videos) while maintaining higher segmentation performance using only 20\% of visual tokens, outperforming all existing pruning baselines.}
  \label{fig:start}
\end{figure}

\section{Introduction}
Instructed Visual Segmentation (IVS) is a task that segments objects in images or videos based on natural language instructions. It includes a variety of applications, such as referring expression segmentation (RES), referring video object segmentation (R-VOS), and more complex reasoning-based tasks. Compared to traditional segmentation with fixed categories, IVS is more challenging as it requires detailed visual understanding and language grounding. MLLMs have become the foundation for IVS due to their strong ability to align vision and language. In images, methods like LISA~\cite{Lai2023LISARS} and PixelLM~\cite{Ren2023PixelLMPR} show competitive performance by using language as a unified interface. For videos, approaches like VISA~\cite{yan2024visa} combine different expert modules but often rely on complicated pipelines and external tools, limiting their efficiency and scalability. To address this, InstructSeg~\cite{wei2024instructseg} was introduced as an end-to-end model that handles both images and videos. However, inference remains slow, especially for video, due to the cost of processing many reference frames. This makes it difficult to deploy in real-world applications.

Visual token pruning has shown promise for improving MLLMs' efficiency in tasks like Visual Question Answering (VQA) and image captioning~\cite{PruMerge,fastv}. Existing methods, such as attention-based heuristics~\cite{PruMerge}, calibration-based methods~\cite{vtw,ye2025fit} and learned compact representations~\cite{m3,li2025tokenpacker}, focus on coarse semantic understanding and overlook the spatial pixel precision needed for segmentation. They do not adapt to the dense token-to-pixel alignment required in IVS, often leading to degraded mask quality. More importantly, we realize visual token pruning for MLLM-based IVS remains largely unexplored, despite its potential to significantly enhance efficiency.

In support of this, we empirically observe that MLLMs used for IVS contain substantial visual token redundancy, highlighting an opportunity for pruning.
Our experiments show that model performance on both image and video tasks tends to saturate when retaining only 60\%–70\% of the visual tokens. This observation motivates us to investigate how different token selection strategies affect segmentation performance. Specifically, we conduct a quantitative analysis using coverage-based metrics, including $\epsilon$-ball (coverage radius) and performance. We observed a consistent trend where more uniform, coverage and representative visual token subset yields higher-quality segmentation.

Building on the observed connection between token coverage and segmentation performance, we propose a novel visual token pruning method tailored for IVS, EVTP-IVS. Our method extends the classic $k$-center algorithm by incorporating spatial information and introducing an adaptive spatial scaling factor, which guides the selection toward subsets that are both coverage-aware and spatially uniform. The method is simple yet effective, requiring no additional supervision or fine-tuning. Furthermore, we provide an information-theoretic analysis to support the theoretical soundness of our design.
Extensive experiments on various video and image based IVS benchmarks demonstrate the effectiveness of our method. It achieves up to 5× inference speed-up on video tasks and 3.5× on image tasks, while retaining comparable performance using only 20\% visual tokens. Our method consistently outperforms existing SOTA pruning methods (see Fig.~\ref{fig:start}).
\textbf{Our main contributions are summarized as follows:}
\begin{itemize}
    \item We conduct the first systematic study of visual token redundancy in MLLMs for IVS, and empirically reveal a strong correlation between token coverage and segmentation performance.
    
    \item We propose a token pruning method based on $k$-center sampling, incorporating spatial information. We further provide an information-theoretic analysis to explain the role of spatial coverage and uniformity in preserving segmentation quality.
    
    \item Our method delivers significantly faster inference with minimal compromise in segmentation quality, outperforming state-of-the-art pruning methods under identical token budgets on standard IVS benchmarks.
\end{itemize}

\section{Related Work}
\subsection{IVS with MLLMs}

Instructed Visual Segmentation (IVS) aims to produce pixel‐level masks from natural‐language prompts, extending traditional Referring Expression Segmentation (RES) to more complex, reasoning‐driven instructions. Early RES methods (e.g.\ CRIS~\cite{wang2022cris}, CGFormer~\cite{tang2023contrastive}) align vision and language via contrastive learning or decoder fusion, while later works (e.g.\ X‑Decoder~\cite{zou2023generalized}, RefTR~\cite{li2021referring}) improve multimodal fusion through task‑specific supervision. Video extensions (e.g.\ R‑VOS, frame‑wise~\cite{bellver2023closer} or mask‑propagation~\cite{cheng2022xmem}) capture temporal continuity but still lack deep language reasoning.

The advent of MLLMs such as Flamingo~\cite{alayrac2022flamingo}, BLIP‑2~\cite{li2023blip} and MiniGPT‑4~\cite{zhu2023minigpt} has spurred efforts to endow them with dense prediction capabilities. By introducing special “segment” tokens or prompt embeddings~\cite{Lai2023LISARS, xia2023gsva, zhang2024psalm} and coupling with segmentation backbones like SAM~\cite{kirillov2023segment} or Mask2Former~\cite{cheng2022masked}, recent studies (e.g.\ PerceptionGPT~\cite{pi2023perceptiongpt}, PixelLMPR~\cite{Ren2023PixelLMPR}) achieve competitive image segmentation. Extensions to video (e.g.\ VISA~\cite{yan2024visa}) interleave frame selection, MLLM reasoning and external memory modules, but remain non‑end‑to‑end and computationally heavy. InstructSeg~\cite{wei2024instructseg} proposes the a unified MLLM framework for both image and video IVS by integrating a vision‑guided fusion module with a video‑aware encoder, which also bring the IVS performances to new level.
Despite its strong performance, its reliance on a large number of token sequences (e.g., many reference frames in video segmentation) leads to prohibitive inference, especially under real‑time or resource‑constrained settings, and to date, efficiency optimization for such unified MLLM segmenters remains largely unexplored.

\subsection{Visual Token Pruning for Efficient Inference}

To improve inference efficiency in MLLMs, visual token pruning has emerged as a promising strategy~\cite{bolya2022token}. Attention-based methods such as FastV~\cite{fastv} and PruMerge~\cite{PruMerge} remove tokens with low attention scores but often retain redundant information. Calibration-based methods like FitPrune~\cite{ye2025fit} and VTW~\cite{vtw} require auxiliary data to optimize pruning schedules, which limits scalability. Fine-tuning-based methods like M3~\cite{m3} and TokenPacker~\cite{li2025tokenpacker}. Recently, inference-time one-shot pruning methods such as DivPrune~\cite{divprune}, VisPruner~\cite{zhang2024beyond}, and VisionZip~\cite{yang2025visionzip} have been proposed to address visual token redundancy by selecting compact yet representative token subsets. These method achieve significant speedups with minimal accuracy loss and can be easily integrated into MLLMs.
\begin{figure*}[htb]
  \centering
  \includegraphics[width=1.0\textwidth]{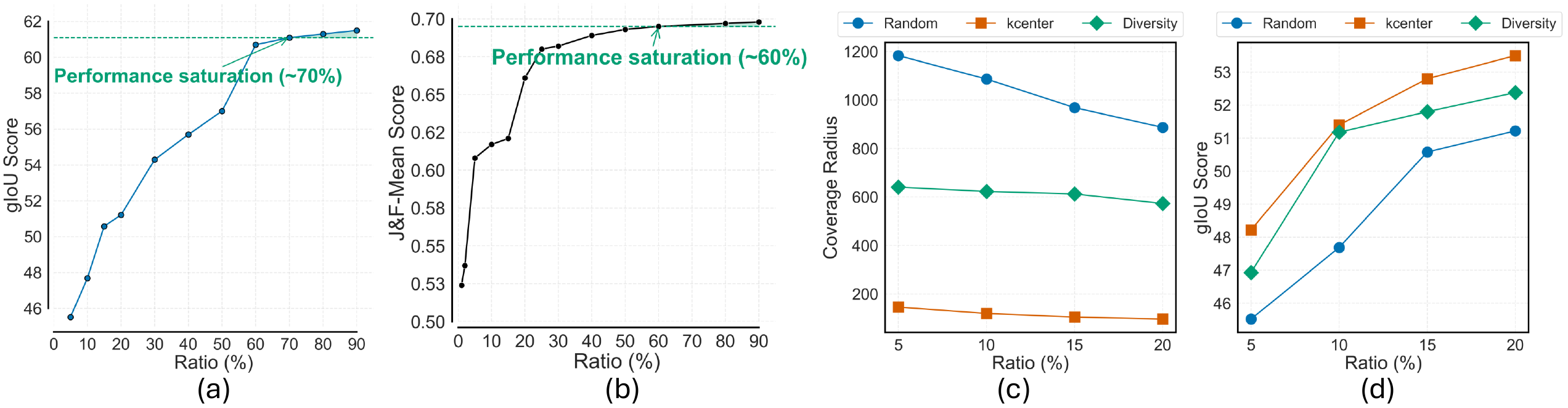}
  \caption{
(a)-(b): Random token sampling performance on image and video segmentation benchmarks, measured by gIoU and J\&F-Mean across varying selected token ratios. Saturation occurs around 60\% and 70\%, respectively. 
(c)-(d): Comparison of coverage radius (lower is better) and performance (gIoU, higher is better) across token selection strategies (Random, $k$-Center, Diversity). The $k$-Center method consistently demonstrates better coverage and improved performance.
}
  \label{fig:pre-experiments}
\end{figure*}
\subsection{Visual Token Pruning in MLLMs based IVS}
However, the majority of existing pruning methods have been developed with high-level vision tasks in mind, such as visual question answering (VQA) and image captioning, where coarse global semantics typically suffice. In contrast, visual segmentation is a mid-level task that demands fine-grained spatial alignment between visual tokens and pixel-level information.
Applying these general pruning strategies often leads to degraded segmentation performance. This gap motivates the need for MLLMs segmentation-aware pruning strategies that explicitly balance performance and efficiency during inference.

\section{Preliminary}

\begin{figure}[t]
  \centering
  \includegraphics[width=0.45\textwidth]{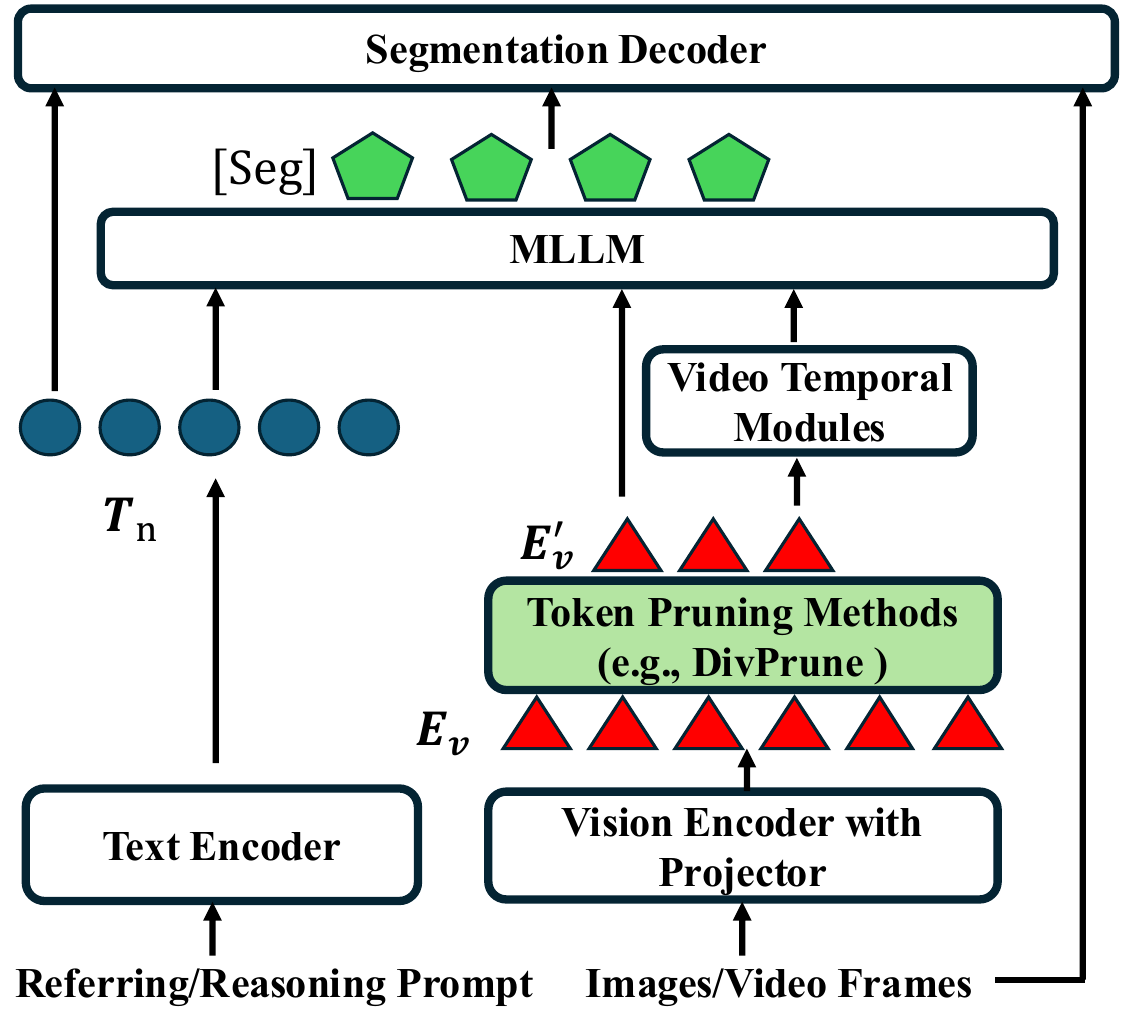}
  \caption{ A schematic of segmentation MLLMs with visual token pruning, where visual tokens are pruned before feeding to the MLLM.}
  \vspace{-0.4cm}
  \label{fig:prune_pre}
\end{figure}

Pruning has never been explored in the context of unifying MLLM-based IVS tasks. Here, we are the first to explicitly define visual token pruning for this setting, aiming to reduce the large number of visual tokens before fusion with language (see Fig.~\ref{fig:prune_pre}).
MLLM-based segmentation models extend large multimodal language models to produce dense, pixel-level predictions from language-guided visual inputs. Given a multimodal input $(T, V)$, where $T$ denotes a natural-language instruction and $V$ denotes a visual signal (an image or a video of $F$ frames), $T$ is encoded into a text token sequence $\mathbf{T}_n = \{t_1,\dots,t_N\}$, and $V$ into a visual token sequence $\mathbf{E}_v = \{v_1,\dots,v_M\}$. For videos, $M = F \cdot P$, where $P$ denotes the number of patches per frame, often resulting in a significant token imbalance ($M \gg N$).
We apply visual token pruning to $\mathbf{E}_v$, selecting a compact subset $\mathcal{C}$ to represent the visual content. This is performed prior to fusion with the language stream, aiming to preserve instruction-relevant semantics while reducing inference-time computation. 
The pruned visual tokens $\mathcal{C}$ are fused with text tokens in an LLM for joint semantic reasoning. Unlike typical MLLMs that produce autoregressive text, it decode the fused representations into segmentation (query) embedding $\mathbf{[Seg]}$. The final mask is obtained by segmentation decoder (e.g., SAM~\cite{kirillov2023segment} or Mask2Former~\cite{cheng2022masked}):
\[
\text{Mask} = \text{Decoder}(f_{\text{img}}, \mathbf{T}_n, \mathbf{[Seg]}),
\]
where $f_{\text{img}}$ denotes visual features (e.g., ViT within segmentation decoder), guided by both text and mask embeddings.
While this formulation supports flexible visual reasoning, the number of visual tokens $M$ can be large, especially for videos IVS with $V_r$ reference frames, tokens scale with $V_r \cdot M$, causing quadratic attention cost and 3--5$\times$ slower inference. Pruning reduces token length before fusion to alleviate this overhead.

\subsection{Visual Token Redundancy} 
To validate whether visual tokens are redundant in both image and video IVS tasks, we conduct empirical studies on InstructSeg~\cite{wei2024instructseg}. Specifically, we randomly subsample a certain percentage of image patch tokens while evaluate the performance.
Fig.~\ref{fig:pre-experiments}(a,b) shows the results on the image and video datasets. In (a), the gIoU score follows a similar trend, plateauing beyond 70\% token ratio. Similarly, in (b), we plot the J\&F-Mean score with respect to different token ratios for video dataset. We observe that the performance saturates at around 60\%. It is evident that visual token redundancy exists in MLLMs based IVS tasks for both images and videos. This observation raises an important question: \textit{what token selection strategies are most effective for preserving performance while minimizing redundancy?}

\subsection{Quantitative Analysis of Token Selection Strategies}
\begin{figure}[t]
  \centering
  \includegraphics[width=0.4\textwidth]{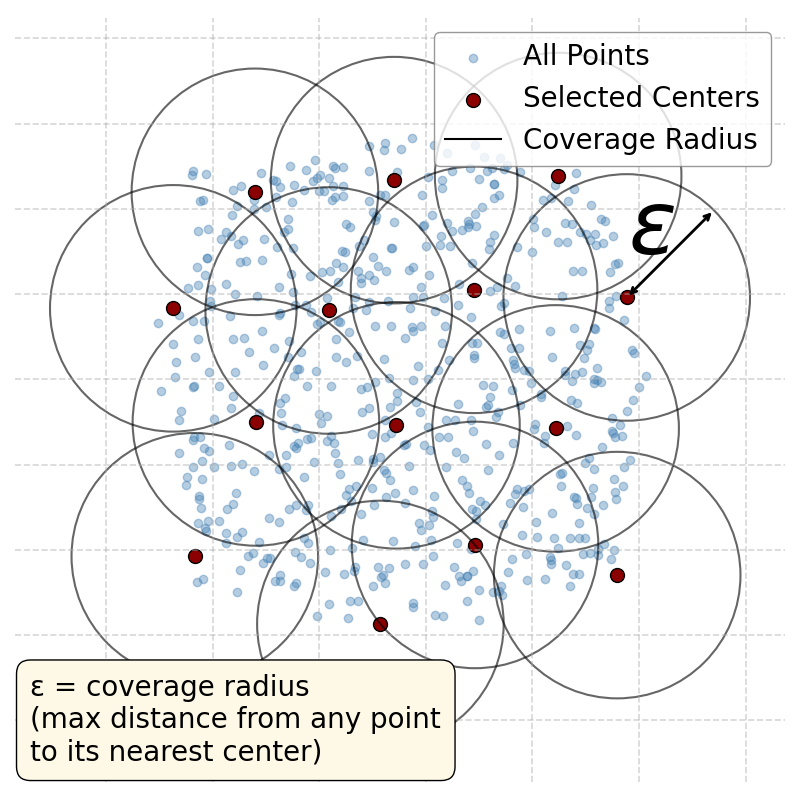}
  \caption{ An conceptual illustration for $\epsilon$-ball and coverage radius.}
  \label{fig:example}
\end{figure}

Our key hypothesis is that better coverage of the input feature space leads to improved mask accuracy, as coverage ensures that both dominant structures and important fine-grained regions are well represented.
Unlike high-level vision-language tasks that often rely on summarizing tokens, segmentation demands a balanced set of tokens that are both diverse and spatially representative to preserve boundaries and context. Hence, to explore how visual token selection impacts segmentation, we start by examining the relationship between token coverage and segmentation quality. We consider three representative token selection strategies: (1) \textit{Random sampling}, as a naive but the most important baseline; (2) \textit{DivPrune \cite{divprune}}, which emphasizes token diversity to reduce redundancy but may sacrifice global coverage; and (3) the \textit{$k$-center algorithm}, which aims to maximize coverage by selecting tokens that best represent the overall feature distribution.
To validate our hypothesis, i.e., better coverage correlates to better performance, we evaluate each method under two complementary criteria: 
\begin{itemize}[leftmargin=*]

\item The \emph{coverage radius}, defined as the maximum distance from any original token $\mathbf{v}$ to its nearest selected token representative $\mathbf{c}$ in a coreset $\mathcal{C}$ as Coverage Radius $ R(\mathcal{C}) = \max_{v \in \mathbf{E}_v} \min_{\mathbf{c} \in \mathcal{C}} \|\mathbf{v} - \mathbf{c}\|_2$. Intuitively, the coverage radius quantifies the worst-case approximation error introduced by the coreset $\mathcal{C}$, or how far the most poorly represented token is from its closest representative in \(\mathcal{C}\). A smaller coverage radius implies better geometric representativeness and uniformity, which is desirable in preserving both spatial and semantic cues for downstream tasks. A conceptual visualization is shown in Fig. \ref{fig:example}. 

\item The resulting \emph{segmentation performance} measured by GIOU.
\end{itemize}

As shown in Figs. \ref{fig:prune_pre}(c)-(d), the primary observation is that at different ratios of selected tokens, $k$-Center achieves the smallest coverage radius among the three methods, indicating more uniform and representative coverage of the visual feature space than both diversity-based method and random selection. More importantly, we find that this coverage successfully translates to segmentation performance improvement, which is consistent with our hypothesis. 

\section{Method}

Based on the empirical observation that better feature coverage, measured by a smaller \emph{coverage radius}, often leads to improved segmentation performances.
However, the vanilla $k$-Center only considers feature similarity and completely ignores the spatial layout of tokens. As a result, it tends to select spatially clustered tokens, which may cause important regions to be missed. To address this, we extend $k$-Center with spatial information by introducing coordinate augmentation and adaptive weighting. This leads to a selection that is not only semantically diverse but also spatially well-distributed.

\noindent \textbf{Feature-space token selection.}  
We begin with the vanilla $k$-Center formulation. Given a set of token embeddings $\mathbf{E}_v = \{v_1, \dots, v_M\}$, the goal is to select a subset $\mathcal{C}$ with $K$ elements that minimizes the maximum distance from each token to its nearest selected token:
\begin{align}
\min_{\mathcal{C} \subset \mathbf{E}_v,\; |\mathcal{C}|=K} \max_{v \in \mathbf{E}_v} \min_{c \in \mathcal{C}} \|v - c\|_2.
\label{eq:kcenter}
\end{align}
This ensures that all tokens are well represented in the feature space, promoting coverage.

\noindent \textbf{Spatial augmentation.}  
To account for spatial structure, we augment each token with its normalized 2D coordinates:
\begin{align}
\mathbf{coords}_i = \left[ \frac{x_i}{W}, \frac{y_i}{H} \right],
\end{align}
where $(x_i, y_i)$ is the token's location on the $W \times H$ grid. We then define the augmented token representation as:
\begin{align}
\widetilde{v}_i = \left[ v_i,\, \lambda\, \mathbf{coords}_i \right],
\end{align}
where $\lambda$ is a scaling factor that determines the contribution of the spatial component. To make the method adaptive to the distribution of features, we set:
\begin{align}
\lambda = \mathrm{Var}(\mathbf{E}_v) + \varepsilon,
\end{align}
where $\mathrm{Var}(\mathbf{E}_v)$ is the variance of the token features and $\varepsilon$ is a small constant for stability. This means that in regions where token features are very similar, such as flat or homogeneous areas, spatial information becomes more important. Before augmentation, all features are normalized:
\begin{align}
\mathbf{E}_v \leftarrow \frac{\mathbf{E}_v - \mu}{\mathrm{Var}(\mathbf{E}_v) + \varepsilon},\quad \mu = \frac{1}{M}\sum_{i=1}^M v_i.
\end{align}

\noindent \textbf{Greedy selection in joint space.}  
After augmenting the features with spatial information, we apply a greedy $k$-Center algorithm in the joint space. The algorithm starts by picking the token that is farthest from the global mean. Then, at each step, it adds the token that has the largest minimum distance to the current selected set:
\begin{align}
s_t = \arg\max_i \min_{j\in S} \| \widetilde{v}_i - \widetilde{v}_j \|_2.
\end{align}

This process continues until $K$ tokens are selected. The complete procedure is shown in Algorithm~\ref{alg:pruning}.
\begin{algorithm}[htp]
\caption{Our Visual Token Pruning}
\label{alg:pruning}
\begin{algorithmic}[1]
  \Require $\mathbf{E}_v = \{v_1,\dots,v_M\},\;r\in(0,1)$  
  \Ensure Pruned tokens $\mathcal{C}$ of size $k=\lfloor rM\rfloor$
  \State Let $\varepsilon>0$ be a small constant
  \State $\mu \gets \frac{1}{M}\sum_{i=1}^M v_i$
    \State $\lambda \gets \mathrm{Var}(\mathbf{E}_v) + \varepsilon$
  \State $\mathbf{E}_v \gets (\mathbf{E}_v - \mu)/(\mathrm{Var}(\mathbf{E}_v) + \varepsilon)$
  \State Compute normalized spatial coords: $\mathbf{coords}_i = [x_i/W,\; y_i/H]$
  \State $\widetilde{v}_i \gets [v_i,\, \lambda \cdot \mathbf{coords}_i]$ for $i=1,\ldots,M$
  \State $k \gets \max(1,\lfloor rM\rfloor)$
  \State $\overline{\widetilde E} \gets \tfrac1M \sum_{i=1}^M \widetilde E_i$
  \State Initialize $S$ with $\arg\max_i \|\widetilde{v}_i - \overline{\widetilde{E}_v}\|_2$ 
  \While{$|S| < k$}
    \State Add $s = \arg\max_i \min_{j\in S} \| \widetilde{v}_i - \widetilde{v}_j \|_2$ to $S$
  \EndWhile
  \State $\mathcal{C} \gets \{v_i: i\in S\}$
  \State \Return $\mathcal{C}$
\end{algorithmic}
\end{algorithm}
The algorithm has a time complexity of $\mathcal{O}(KM)$ and is efficient even for large token sets. The pruning ratio $r$ provides a flexible control over the trade-off between computational cost and segmentation performance.

\section{Theoretical Analysis}

We show that jointly optimizing features and spatial coordinates enhances token selection from information-theoretic analysis.

\begin{definition}[Feature Coverage Radius]
Let $\mathbf{E}_v = \{v_1, \ldots, v_M\} \subset \mathbb{R}^D$ be the set of visual feature tokens. For any subset $\mathcal{C} \subset \mathbf{E}_v$, define its \emph{feature coverage radius} as
\[
R_f(\mathcal{C}) = \max_{v \in \mathbf{E}_v} \min_{c \in \mathcal{C}} \|v - c\|_2.
\]
\end{definition}

\begin{definition}[Joint Coverage Radius]
Augment each token $v_i$ with its normalized spatial coordinate $(x_i, y_i) \in [0,1]^2$ and a weight parameter $\lambda > 0$: 
\[
\widetilde{x}_i = (v_i, \lambda(x_i, y_i)) \in \mathbb{R}^{D+2}.
\]
 For any subset $\mathcal{C} \subset \mathbf{E}_j  = \{\widetilde{x}_1, \ldots, \widetilde{x}_M\}$, define its \emph{joint coverage radius} as
\[
R_j(\mathcal{C}) = \max_{\widetilde{x}_i \in \mathbf{E}_j} \min_{\widetilde{c} \in \mathcal{C}} \|\widetilde{x}_i - \widetilde{c}\|_2.
\]
\end{definition}

\begin{definition}[Spatial Coverage Radius]
For any subset $\mathcal{C} \subset \{(x_1, y_1), \ldots, (x_M, y_M)\}$, define its \emph{spatial coverage radius} as
\[
R_s(\mathcal{C}
) = \max_{(x_i, y_i)} \min_{(x_c, y_c) \in \mathcal{C}} \|(x_i, y_i) - (x_c, y_c)\|_2.
\]
\end{definition}

We now connect coverage radius to segmentation quality from Information-theoretic perspective.

\begin{assumption}
Let $I(\mathcal{C}; \mathbf{E}_x)$ denote the learning quality (e.g., representational or predictive power) achieved by selecting a token subset $\mathcal{C}$ from an embedding space $\mathbf{E}_x$. This reflects the common assumption that mutual information correlates with segmentation performance, as supported by prior work such as~\cite{tishby2015deep, alemi2016deep, cover2006elements, chen2025sequence}.
\end{assumption}

\begin{theorem}
Let $\mathcal{C}_v^* = \arg\max_{\mathcal{C} \subset \mathbf{E}_v,\ |\mathcal{C}| = k} I(\mathcal{C}; \mathbf{E}_v)$ and $C_j^* = \arg\max_{\mathcal{C} \subset \mathbf{E}_j,\ |\mathcal{C}| = k} I(\mathcal{C}; \mathbf{E}_j)$ denote the optimal $k$-Center selections from the feature space $\mathbf{E}_v$ and joint space $\mathbf{E}_j$, respectively. 
Then the following inequality holds:
\[
I(\mathcal{C}_j^*; \mathbf{E}_j) \ge I(\mathcal{C}_v^*; \mathbf{E}_v)
\]
\end{theorem}

\begin{proof}

Let $f: \mathbf{E}_j \rightarrow \mathbf{E}_v$ be a deterministic projection that discards spatial coordinates. For any subset $\tilde{\mathcal{C}}_v \subset \mathbf{E}_j$ such that $f(\tilde{\mathcal{C}}_v) = \mathcal{C}_v$, the data processing inequality~\cite{cover2006elements} implies:
\[
I(\mathcal{C}_v; \mathbf{E}_v) = I(f(\tilde{\mathcal{C}}_v); \mathbf{E}_v) \le I(\tilde{\mathcal{C}}_v; \mathbf{E}_j)
\]

In particular, for the optimal $\mathcal{C}_v^*$, we have:
\[
\begin{aligned}
  I(\mathcal{C}_v^*; \mathbf{E}_v)
    &\le I(\tilde{\mathcal{C}}_v^*; \mathbf{E}_j) \\
    &\le \max_{\substack{\mathcal{C}\subset \mathbf{E}_j \\ |\mathcal{C}|=k}} I(\mathcal{C}; \mathbf{E}_j) \\
    &= I(\mathcal{C}_j^*; \mathbf{E}_j).
\end{aligned}
\]

Therefore,
\[
I(\mathcal{C}_v^*; \mathbf{E}_v) \le I(\mathcal{C}_j^*; \mathbf{E}_j)
\]
as claimed.
\end{proof}

\begin{remark}
The theorem immediately suggests that the optimal selection from $\mathbf{E}_j$ can preserve more information relevant to the input, which may translate to improved learning performance in downstream tasks.     
\end{remark}

\begin{remark}
Spatial augmentation helps avoid redundant selections from the same local region. By spreading the selected tokens more evenly across the image, we indirectly improve feature-space coverage as well. The parameter $\lambda$ controls how strongly spatial diversity influences the final selection and is adapted automatically based on the distribution of features.
\end{remark}

\begin{table*}[t]
  \centering
  \caption{Performance comparison on referring expression segmentation (refCOCO, refCOCO+, refCOCOg) and ReasonSeg. 
The  Ratio specifies the ratio of visual tokens used. 
 \textbf{Segmentation visualization analysis is provided in Appendix~B.}}
  \resizebox{1\textwidth}{!}{
    \begin{tabular}{c|c|c|ccc|ccc|cc|cc}
    \toprule[1.1pt] 
    \multirow{2}{*}{Ratio} & \multirow{2}{*}{Method} & \multirow{2}{*}{TFLOP} & \multicolumn{3}{c|}{refCOCO} & \multicolumn{3}{c|}{refCOCO+} & \multicolumn{2}{c|}{refCOCOg} & \multicolumn{2}{c}{ReasonSeg} \\
    \cline{4-13}
    & & & val & testA & testB & val & testA & testB & val(U) & test(U) & gIoU & cIoU \\
    \midrule

    \cellcolor[gray]{0.85} \textbf{100\%} & \cellcolor[gray]{0.85} Original & \cellcolor[gray]{0.85} \textbf{2.376} & \cellcolor[gray]{0.85} \textbf{85.1} & \cellcolor[gray]{0.85} \textbf{85.8} & \cellcolor[gray]{0.85} \textbf{83.7} & \cellcolor[gray]{0.85} \textbf{81.1} & \cellcolor[gray]{0.85} \textbf{84.1} & \cellcolor[gray]{0.85} \textbf{78.1} & \cellcolor[gray]{0.85} \textbf{79.6} & \cellcolor[gray]{0.85} \textbf{80.3} & \cellcolor[gray]{0.85} \textbf{61.9} & \cellcolor[gray]{0.85} \textbf{65.2} \\

    \midrule
    \multirow{5}{*}{\textbf{5\%}}

    & ToM (ICML 23) &     & \textbf{71.7} & \textbf{72.5} & \textbf{72.4} & 54.6 & 58.1 & 52.1  & 58.6 & 58.8 & 43.2 & 43.8   \\
    & VisionZip (CVPR 25) &     & 67.8 & 68.9 & 68.5 & 47.9 & 50.6 & 47.3  & 50.6 & 51.0 & 39.4 & 46.1 \\
    & VisPruner (ICCV 25) & 0.447   & 67.7 & 66.8 & 68.0 & 50.3 & 52.2 & 49.9  & 51.5 & 52.2 & 39.8 & 42.8 \\
    &  DivPrune (CVPR 25) &  &  69.9 &  70.9 &  70.1 &  \textbf{55.3} &  57.9 &  \textbf{54.9} &  58.7 &  59.1 &  46.9 &  51.4\\
    & \textbf{Our} &  & 68.9 & 69.0 & 69.0 & 54.8 & \textbf{58.2} & 52.6 & \textbf{58.9} & \textbf{59.3} & \textbf{47.5} & \textbf{53.4} \\

    \midrule
    \multirow{5}{*}{\textbf{10\%}}

    & ToM (ICML 23) &     & 74.6 & 76.0 & 74.4 & 60.2 & 64.5 & 56.9  & 62.3 & 64.0 & 48.0 & 53.6 \\
    & VisionZip (CVPR 25) &     & 70.6 & 71.0 & 69.7 & 52.1 & 56.0 & 49.8  & 54.7 & 55.0 & 42.6 & 47.7 \\
    & VisPruner (ICCV 25) & 0.525   & 70.8 & 72.0 & 71.4 & 55.7 & 59.7 & 53.0  & 58.5 & 58.9 & 44.8 &  49.3 \\
    &  DivPrune (CVPR 25) &  &  \textbf{75.6} &  76.9 &  74.5 &  65.2 &  \textbf{69.0} &  \textbf{62.4} &  67.5 &  68.4 &  50.9 &  52.6 \\
    & \textbf{Our} &  & 75.4 & \textbf{77.2} & \textbf{74.8} & \textbf{66.0} & 68.8 & 61.5 & \textbf{68.7} & \textbf{68.8} & \textbf{51.3} & \textbf{55.6} \\

    \midrule
    \multirow{5}{*}{\textbf{20\%}}

    & ToM (ICML 23) &     & 76.2 & 78.0 & 76.6 & 64.6 & 69.2 & 60.2  & 66.5 & 67.3 & 48.9 & 52.1 \\
    & VisionZip (CVPR 25) &     & 70.7 & 72.0 & 70.2 & 55.0 & 58.9 & 52.4  & 56.0 & 56.9 & 46.3 & 49.5 \\
    & VisPruner (ICCV 25) & 0.724   & 72.4 & 72.7 & 72.7 & 58.1 & 61.9 & 56.3  & 60.5 & 62.2 & 49.0 & 52.4 \\
    &  DivPrune (CVPR 25) &  &  80.2 &  81.0 &  78.6 &  72.0 &  76.6 &  68.3 &  73.6 &  73.9 &  52.3 &  53.0  \\
    & \textbf{Our} &  & \textbf{80.3} & \textbf{81.8} & \textbf{78.6} & \textbf{73.0} & \textbf{77.3} & \textbf{68.9} & \textbf{74.7} & \textbf{74.3} & \textbf{54.3} & \textbf{56.1} \\

    
    \bottomrule[1.1pt]
    \end{tabular}
  }
  \label{tab:seg_compare}
\end{table*}

\section{Experiments}

\subsection{Baselines and Model}

We use InstructSeg~\cite{wei2024instructseg} as the base model. As a recent state-of-the-art model, it offers a unified framework for both image and video segmentation, making it well-suited for evaluating general-purpose pruning strategies. For fair comparison, we focus on SOTA general one-shot pruning methods including ToM~\cite{bolya2022token}, VisionZip~\cite{yang2025visionzip}, VisPruner~\cite{zhang2024beyond}, and DivPrune~\cite{divprune} without fine-tuning; these methods are compatible with InstructSeg. We exclude layer-wise pruning and fine-tuning-based methods, as they are either not directly applicable to temporal modules (e.g., FastV~\cite{fastv}, VTW~\cite{vtw}); or require retraining, e.g. as in TokenPacker~\cite{li2025tokenpacker} and M3~\cite{m3}.

\begin{table*}[t]
  \centering
  \caption{Comparison on the ReVOS benchmark (reasoning video object segmentation). 
Our method consistently outperforms all pruning methods across different token retention ratios (``Ratio'' indicates the percentage of visual tokens retained after pruning). 
\textbf{Segmentation visualizations Analysis is provided in Appendix~B.}}
  \resizebox{1\textwidth}{!}{
    \begin{tabular}{c|c|c|ccc|ccc|ccc}
    \toprule[1.1pt] 
    \multirow{2}{*}{Ratio} & \multirow{2}{*}{Method} & \multirow{2}{*}{TFLOP} & \multicolumn{3}{c|}{Reasoning} & \multicolumn{3}{c|}{Referring} & \multicolumn{3}{c}{Overall} \\
    \cline{4-12}
    & & & $\mathcal{J}$ & $\mathcal{F}$ &  $\mathcal{J\&F}$ & $\mathcal{J}$ & $\mathcal{F}$ &  $\mathcal{J\&F}$ & $\mathcal{J}$ & $\mathcal{F}$ &  $\mathcal{J\&F}$ \\
    \midrule[0.7pt]
    
    \cellcolor[gray]{0.85} \textbf{100\%} & \cellcolor[gray]{0.85} \textbf{Original} & \cellcolor[gray]{0.85} \textbf{9.609} & \cellcolor[gray]{0.85} \textbf{49.2} & \cellcolor[gray]{0.85} \textbf{54.7} & \cellcolor[gray]{0.85} \textbf{51.9} & \cellcolor[gray]{0.85} \textbf{54.8} & \cellcolor[gray]{0.85} \textbf{59.2} & \cellcolor[gray]{0.85} \textbf{57.0} & \cellcolor[gray]{0.85} \textbf{52.0} & \cellcolor[gray]{0.85} \textbf{56.9}  & \cellcolor[gray]{0.85} \textbf{54.5} \\

    \midrule

    \multirow{5}{*}{\textbf{5\%}}

    & ToM (ICML 23)  &  & 36.4 & 42.0 & 39.2 & 40.9 & 46.4 & 43.7 & 38.6 & 44.2 & 41.4 \\
    & VisionZip (CVPR 25)  &  & 32.8 & 38.8 & 35.8 & 36.7 & 42.7 & 39.7 & 34.8 & 40.8 & 37.8 \\
    & VisPruner (ICCV 25)  & 0.751 & 35.9 & 41.8 & 38.9 & 42.1 & 47.5 & 44.8 & 39.0 & 44.7 & 41.8 \\
    &  DivPrune (CVPR 25)  &  & 40.6 & 46.3 & 43.5 & 44.2 & 49.6 & 46.9 & 42.4 & 47.9 & 45.2 \\
    & \textbf{Our} &  & \textbf{40.6} & \textbf{46.4} & \textbf{43.5} & \textbf{45.6} & \textbf{50.7} & \textbf{48.0} & \textbf{43.0} & \textbf{48.5} & \textbf{45.8}  \\

    \midrule

    \multirow{5}{*}{\textbf{10\%}}

    & ToM (ICML 23)  &  & 38.8 & 44.4 & 41.6 & 43.6 & 49.0 & 46.3 & 41.2 & 46.7 & 43.9 \\
    & VisionZip (CVPR 25)  &  & 34.5 & 40.6 & 37.6 & 39.0 & 44.9 & 42.0 & 36.8 & 42.7 & 39.8 \\
    & VisPruner (ICCV 25)  & 1.161 & 38.1 & 43.8 & 40.9 & 44.7 & 49.9 & 47.3 & 41.4 & 46.8 & 44.1 \\
    &  DivPrune (CVPR 25)  &  & 43.1 & 48.7 & 45.9 & 48.6 & 53.6 & 51.1 & 45.9 & 51.1 & 48.5 \\
    & \textbf{Our} &  & \textbf{43.6} & \textbf{49.3} & \textbf{46.4} & \textbf{49.7} & \textbf{54.7} & \textbf{52.2} & \textbf{46.6} & \textbf{52.0} & \textbf{49.3}  \\

    \midrule
        \multirow{5}{*}{\textbf{20\%}}

    & ToM (ICML 23)  &  & 39.7 & 45.3 & 42.5 & 45.4 & 50.7 & 48.1 & 42.6 & 48.0 & 45.3 \\
    & VisionZip (CVPR 25) &  & 35.7 & 41.6 & 38.6 & 40.5 & 46.3 & 43.4 & 38.1 & 44.0 & 41.0 \\
    & VisPruner (ICCV 25)  & 1.989 & 40.0 & 44.6 & 41.7 & 45.8 & 50.9 & 48.4 & 42.4 & 47.7 & 45.1 \\
    &  DivPrune (CVPR 25)  &   & 45.0 & 50.5 & 47.7 & 50.9 & 55.7 & 53.3 & 48.0 & 53.1 & 50.5 \\
    & \textbf{Our} &  & \textbf{45.4} & \textbf{51.1} & \textbf{48.2} & \textbf{51.9} & \textbf{56.7} & \textbf{54.3} & \textbf{48.7} & \textbf{53.9} & \textbf{51.3}  \\


        
    \bottomrule[1.1pt]
    \end{tabular}
  }
  \label{tab:reason_revos}
\end{table*}

\begin{table*}[h]
  \centering
  \caption{Results on referring video object segmentation benchmarks: Ref-YouTube-VOS (val) and Ref-DAVIS17 (val). Our method achieves superior performance across all ratios. \textbf{Segmentation visualizations analysis is provided in Appendix~B.}}
  \resizebox{0.8\textwidth}{!}{
    \begin{tabular}{l|l|c|ccc|ccc}
    \toprule[1.1pt] 
    \multirow{2}{*}{Ratio} & \multirow{2}{*}{Method} & \multirow{2}{*}{TFLOP} & \multicolumn{3}{c|}{Ref-YouTube-VOS (val)} & \multicolumn{3}{c}{Ref-DAVIS17 (val)} \\
    \cline{4-9}
    & & & $\mathcal{J}$ & $\mathcal{F}$ & $\mathcal{J\&F}$ & $\mathcal{J}$ & $\mathcal{F}$ & $\mathcal{J\&F}$ \\
    \midrule[0.7pt]

    \cellcolor[gray]{0.85} \textbf{100\%} & \cellcolor[gray]{0.85} \textbf{Original } & \cellcolor[gray]{0.85} \textbf{9.590} & \cellcolor[gray]{0.85} \textbf{65.4} & \cellcolor[gray]{0.85} \textbf{69.5} & \cellcolor[gray]{0.85} \textbf{67.5} & \cellcolor[gray]{0.85} \textbf{67.3} & \cellcolor[gray]{0.85} \textbf{74.9} & \cellcolor[gray]{0.85} \textbf{71.1} \\

    \midrule
    \multirow{5}{*}{\textbf{5\%}}

    & ToM (ICML 23)  &   & 51.4 & 55.7 & 53.6 & 54.0 & 62.7 & 58.3  \\
    & VisionZip (CVPR 25)  &  & 45.0 & 49.9 & 47.5 & 49.3 & 53.4 & 53.5  \\
    & VisPruner (ICCV 25)  & 0.738 & 51.6 & 56.0 & 53.8 & 53.8 & 62.5 & 58.2  \\
    &  DivPrune (CVPR 25)  &  & 55.4 & 59.3 & 57.3 & 56.5 & 65.0 & 60.7  \\
    & \textbf{Our} &   & \textbf{56.2} & \textbf{60.3} & \textbf{58.2} & \textbf{57.5} & \textbf{65.7} & \textbf{61.6}  \\

    \midrule

    \multirow{5}{*}{\textbf{10\%}}

    & ToM (ICML 23)  &   & 55.0 & 59.2 & 57.1 & 58.0 & 66.2 & 62.1  \\
    & VisionZip (CVPR 25) &  & 49.2 & 53.8 & 51.5 & 52.6 & 60.7 & 56.7  \\
    & VisPruner (ICCV 25)  & 1.147 & 55.0 & 57.1 & 59.2 & 56.4 & 65.2 & 60.8  \\
    &  DivPrune (CVPR 25)  &  & 59.4 & 63.4 & 61.4 & 59.2 & 67.6 & 63.4  \\
    & \textbf{Our} &   & \textbf{60.9} & \textbf{65.0} & \textbf{63.0} & \textbf{60.5} & \textbf{68.5} & \textbf{64.5}  \\

    \midrule
        \multirow{5}{*}{\textbf{20\%}}

    & ToM (ICML 23)  &  & 56.8 & 61.0 & 58.9 & 58.6 & 66.8 & 62.7  \\
    & VisionZip (CVPR 25)  &  & 51.7 & 56.1 & 53.9 & 54.3 & 62.3 & 58.3  \\
    & VisPruner (ICCV 25)  & 1.975 & 56.4 & 60.6 & 58.5 & 55.6 & 64.4 & 60.0  \\
    &  DivPrune (CVPR 25)  &  & 62.2 & 66.1 & 64.2 & 59.7 & 68.0 & 63.9  \\
    & \textbf{Our} &   & \textbf{63.2} & \textbf{67.2} & \textbf{65.2} & \textbf{60.6} & \textbf{68.8} & \textbf{64.7}  \\


    
    \bottomrule[1.1pt]
    \end{tabular}
  }
  \label{tab:exp-r-vos}
\end{table*}

\subsection{Datasets and Metric}
For referring‐expression segmentation, we employ the RefCOCO, RefCOCO+ and RefCOCOg benchmarks~\cite{yu2016modeling,nagaraja2016modeling} to evaluate image‐level performance, and Ref-DAVIS17 and Ref‑YouTube‑VOS~\cite{seo2020urvos} for video‐based referring segmentation. For reasoning segmentation tasks, we utilize the ReasonSeg dataset~\cite{Lai2023LISARS} to assess image‐level inference segmentation and the ReVOS benchmark~\cite{yan2024visa} to evaluate video‐level reasoning segmentation.
We evaluate referring expression image segmentation using global Intersection‐over‐Union (gIoU). For image reasoning segmentation, we report both cumulative IoU (cIoU) and gIoU. Video‐level segmentation performance is measured by region similarity \(\mathcal{J}\) and contour accuracy \(\mathcal{F}\).
Following~\cite{divprune,vtw,fastv}, the total compute (in TFLOPs) is given by
\[
\mathrm{TFLOPs}
= \frac{S \;+\; N\bigl(4\mu\,d^2 \;-\;2\mu^2 d \;+\;2\mu\,d\,D\bigr)}{10^{12}},
\]
where \(\mu = T_{\mathrm{text}} + M + L\), $T_{\mathrm{text}}$ is the number of text tokens, \(S\) is the FLOPs of shared components, \(N\) the number of transformer layers, \(d\) the hidden size, \(D\) the FFN size, \(M\) the number of visual tokens, and \(L\) the fixed extra token count (e.g. SEG andSYS). For the pruned model, \(\mu\) is replaced by \(\mu' = T_{\mathrm{text}} + M' + L\), where $M'$ denotes the pruned visual tokens. Different with~\cite{divprune}, which estimates FLOPs analytically using approximated decoder operations, our computation explicitly aggregates the FLOPs of each module, i.e., vision encoder, LLM decoder, mask decoder, temporal processing, and VMTF fusion based on~\cite{wei2024lasagna}. This provides a more faithful measurement of end-to-end compute cost. \textbf{More details are in Appendix A.}

\subsection{Image-based IVS Tasks}

We evaluate our method on four image-based IVS benchmarks: refCOCO, refCOCO+, refCOCOg, and ReasonSeg. Unlike typical visual understanding tasks, segmentation in MLLMs requires fine-grained region-level reasoning, making token pruning significantly more challenging. The results are shown in Table~\ref{tab:seg_compare}.

\textbf{refCOCO / refCOCO+ / refCOCOg.}
At 5\% token retention, our method performs slightly below the best methods (e.g., ToM and DivPrune) on refCOCO. However, on refCOCO+ and refCOCOg, which demand stronger generalization and reasoning, our method consistently outperforms all baselines. At 20\% pruning, we achieve 74.3\% on refCOCOg (test), setting a new state-of-the-art under compression. 

\textbf{ReasonSeg.}
On ReasonSeg, which is designed to test higher-order reasoning ability, our method shows a clear advantage for all pruning ratios. Even at 5\%, we achieve 47.5\% gIoU and 53.4\% cIoU, outperforming all compared methods. At 20\%, we reach 56.1\% cIoU, confirming the effectiveness of our method in preserving semantic reasoning under limited token budgets.

\textbf{Efficiency.}
Our method also offers a significant reduction in inference cost. Compared to the original model (2.376 TFLOPs), we reduce computation by up to \textbf{5.3$\times$}. Among all methods, ours is the \textbf{only one} that consistently achieves top performance with the \emph{lowest inference cost} across most benchmarks. This makes it a strong choice for deploying MLLMs in practical, resource-constrained scenarios.

\subsection{Video-based IVS Tasks}

We further evaluate our method on video-based instructed visual segmentation (IVS) tasks, including ReVOS, Ref-YouTube-VOS, and Ref-DAVIS17. These benchmarks require models to handle not only spatial reasoning but also temporal consistency and instruction grounding across frames, posing greater challenges than image segmentation.

\textbf{ReVOS.}
Table~\ref{tab:reason_revos} shows the results. Across all pruning levels, our method achieves the best performance in both reasoning and referring segments. At 10\% token retention, we achieve an overall $\mathcal{J\&F}$ score of 49.3, slightly surpassing DivPrune (48.5) and notably outperforming VisPruner (44.1) and ToM (43.9). Even at 5\%, our performance matches DivPrune on reasoning, and exceeds all baselines on referring and overall scores. This suggests our method preserves key spatiotemporal reasoning capabilities even under tight token budgets.

\textbf{Ref-YouTube-VOS and Ref-DAVIS17.}
As shown in Table~\ref{tab:exp-r-vos}, our method achieves consistently strong results across both video referring segmentation datasets. At 20\% pruning, we reach 65.2 on Ref-YouTube-VOS and 64.7 on Ref-DAVIS17 ($\mathcal{J\&F}$), outperforming the SOTA method (DivPrune) by a margin. Even under 5\% pruning, our model maintains competitive accuracy, suggesting robustness to heavy compression.

\textbf{Efficiency.}
In addition to accuracy, these pruning methods are highly efficient. While the original model runs at 9.6 TFLOPs, these methods achieve competitive performance at a fraction of the cost (e.g., 0.75 TFLOPs with 5\% token), reducing inference cost by up to 12.8$\times$. Our method balances speed and accuracy, and generalizes well to longer, instruction-guided video inputs.

\section{Ablation Analysis}

It is worth mentioning that our method is fully deterministic and does not require any hyperparameter tuning. To evaluate the role of each component, we begin with the basic \textit{k}-Center and progressively add spatial and adaptive features. As reported in Table~\ref{tab:ablation_kcenter}, incorporating spatial coordinates leads to significant improvements on both the ReasonSeg and ReVOS benchmarks. We then introduce an adaptive scaling factor, followed by a deterministic initialization strategy that selects the token farthest from the global feature mean. This design achieves the best performance, reaching 54.3 gIoU on ReasonSeg and 51.3 $\mathcal{J\&F}$ on ReVOS.  The benefit of adding spatial information also aligns with our theoretical analysis to preserve spatial coverage during visual token pruning.

\begin{table}[t]
  \centering
  \caption{
Ablation study by progressively adding components to \textit{k}-center under 20\% token retention.
gIoU is evaluated on the ReasonSeg (image) dataset, and $\mathcal{J\&F}$ on the ReVOS (video) dataset.
}
\resizebox{1\linewidth}{!}{
\begin{tabular}{l|c|c}
\toprule
\textbf{Method} & \textbf{ReasonSeg (gIoU ↑)} & \textbf{ReVOS ($\mathcal{J\&F}$ ↑)} \\
\midrule
\textit{k}-center & 53.5 & 50.5 \\
+ Spatial Coordinate & 54.0 & 51.0 \\
+ Adaptive Scaling Factor & 54.2 & 51.1 \\
\rowcolor[gray]{0.93}
+ Initialization (Ours) & \textbf{54.3} & \textbf{51.3} \\
\bottomrule
\end{tabular}
}
\label{tab:ablation_kcenter}
\end{table}

\section{Conclusion}
In this work, we present the first study on visual token pruning for IVS. We make two key observations: (1) IVS based MLLMs exhibit visual token redundancy, and (2) segmentation performance is strongly correlated with the coverage. These findings suggest that effective pruning should focus not only on diversity but also on coverage.
To this end, we propose EVTP-IVS, a simple yet effective pruning that selects spatially representative and coverage tokens and also achieves significant inference acceleration while maintaining strong segmentation performance. We believe this work provides new insights into efficient IVS and lays the foundation for future research on IVS token pruning in MLLMs.

\bibliographystyle{plainnat} 
\bibliography{bibmain}

\newpage
\appendix
\section{Implementation Details}

\subsection{Baseline Model: \textit{InstructSeg}}
\label{subsec:baseline_model}

We adopt \textit{InstructSeg}~\cite{wei2024instructseg} as the baseline model throughout our experiments. It is a modular multi-modal architecture designed for instruction-driven segmentation tasks over both images and videos, supporting various scenarios such as referring expression segmentation and reasoning-based segmentation. The model consists of a set of fixed pre-trained components, each responsible for a distinct stage of visual-language processing, including visual encoding, token fusion, temporal reasoning, and mask decoding. No modules are fine-tuned during inference.

\paragraph{Language Model.}  
The core reasoning component is a 2.7B-parameter \textsc{Phi-2}~\cite{javaheripi2023phi} language model. It receives tokenized inputs including text instructions, visual tokens, reference frame tokens (for videos), and learnable mask tokens. These are concatenated and processed jointly to produce contextualized embeddings that guide the subsequent segmentation modules. The language model is adapted using Low-Rank Adaptation (LoRA) with a rank of 8. We directly use the publicly released LoRA-tuned weights without further training.

\paragraph{Visual Encoders and Token Pruning.}  
\textit{InstructSeg} employs two vision backbones. First, a \textsc{SigLIP} encoder~\cite{zhai2023sigmoid} is used to extract global visual tokens from each input image or frame. These tokens are aligned with text tokens through the language model. To reduce computational overhead and redundancy, we apply several token pruning methods directly to the SigLIP output. This approach retains important semantic information while reducing the number of visual tokens passed into the language model, leading to more efficient inference.
In addition, a \textsc{Swin-B} transformer~\cite{liu2021swin} is used to extract high-resolution spatial features. These features are not pruned and are passed directly to the segmentation decoder to support fine-grained mask prediction.

\paragraph{Segmentation Decoder.}  
For generating pixel-level predictions, we use a pre-trained \textsc{Mask2Former}~\cite{cheng2022masked} decoder. It combines a pixel decoder with a transformer-based instance decoder to predict $N$ mask proposals. Each proposal is associated with an embedding and a confidence score. The most relevant masks are selected by computing the similarity between predicted mask embeddings and instruction-guided features.

\paragraph{Object-aware Video Perceiver.}  
To support temporal reasoning in video tasks, the model includes an Object-aware Video Perceiver (OVP) module. This component processes a sequence of $T_r = 4$ reference frames alongside the instruction tokens. Cross-attention and $N_1 = 3$ transformer layers are used to extract compressed reference tokens that encode object-specific dynamics over time. These tokens are integrated into the language model to provide temporal context for segmentation.

\paragraph{Vision-guided Text Fusion.}  
To better connect the instruction semantics with the visual scene, a Vision-guided Multi-granularity Text Fusion (VMTF) module is used after the language model. It fuses coarse and fine-grained instruction embeddings with spatial visual features using $N_2 = 3$ cross-attention layers. The resulting embeddings act as dynamic classifiers for the segmentation decoder.

\paragraph{Inference Pipeline.}  
During inference, the model receives an image or key frame $\mathcal{V} \in \mathbb{R}^{H \times W \times 3}$, an optional reference sequence $\mathcal{V}_r$ (for videos), and a natural language instruction $\mathcal{E}$. Visual and reference tokens are extracted and pruned (when applicable), then combined with instruction and mask tokens and passed through the language model. Refined embeddings from VMTF are used to produce final segmentation masks. All components are pre-trained and remain fixed. Architectural hyperparameters such as the number of queries or attention layers follow the default configuration in the original implementation.

\vspace{1mm}
InstructSeg provides a strong and flexible baseline for instruction-conditioned segmentation. Its modular design and compatibility with pruning make it a suitable testbed for analyzing token efficiency in visual-language models.

\subsection{TFLOPs Estimation Setup}
\label{subsec:tflops-setup}

To estimate computational cost, we implement a FLOPs calculator that evaluates the number of floating-point operations (FLOPs) for each major component in InstructSeg. We measure total TFLOPs (in units of $10^{12}$ FLOPs) for a single forward pass. The analysis includes the language model, visual encoder, token pruning module, mask decoder, temporal reasoning layers, and the fusion module (VMTF).

\paragraph{Dataset Inputs.}
We evaluate FLOPs on both image-based and video-based segmentation datasets. For each dataset, we define the number of text tokens $T$, visual tokens per frame $V$, and the number of frames $F$. Table~\ref{tab:tflops-dataset} summarizes the token configuration for each benchmark:

\begin{table}[h]
\centering
\caption{Input configurations for FLOPs estimation.}
\label{tab:tflops-dataset}
\resizebox{0.45\textwidth}{!}{%
\begin{tabular}{lccc}
\toprule
\textbf{Dataset} & \textbf{Text Tokens ($T$)} & \textbf{Visual Tokens ($V$)} & \textbf{Frames ($F$)} \\
\midrule
RefCOCO     & 15  & 729 & 1 \\
RefCOCO+    & 15  & 729 & 1 \\
RefCOCOg    & 23  & 729 & 1 \\
MM-Conv     & 50  & 729 & 1 \\
ReasonSeg   & 80  & 729 & 1 \\
RefYouTube  & 20  & 729 & 4 \\
RefDAVIS    & 18  & 729 & 4 \\
ReVOS       & 25  & 729 & 4 \\
\bottomrule
\end{tabular}%
}
\end{table}

\paragraph{Token Pruning and Fixed Tokens.}
To evaluate the impact of token pruning, we test four pruning settings with $V' \in \{729, 146, 73, 36\}$ visual tokens retained per frame, corresponding to pruning ratios from 0\% to 95\%. For video data, the total number of visual tokens is $V' \times F$. Each input also includes a fixed set of $T_\mathrm{fixed} = 100$ auxiliary tokens (e.g., prompts, [MASK], [CLS]), resulting in a total sequence length:
\[
S = T + T_\mathrm{fixed} + V'.
\]

\begin{table*}[t]
    \centering
    \caption{Task-specific language instructions used for each instructed segmentation task.}
    \label{tab:prompt_design}
    \renewcommand{\arraystretch}{1.6}
    \begin{tabularx}{\textwidth}{l|c|c|X|X}
        \toprule
        \textbf{Task} & \textbf{Visual Type} & \textbf{Dataset} & \textbf{Instruction Template} & \textbf{Example Text Prompt} \\
        \midrule
        
        Referring Expression Segmentation & Image & RefCOCO / + / g &
        \textit{You need to perform Referring Expression Segmentation on the image according to the Text Prompt.} &
        \textit{"A baseball catcher with an open mitt"} \\

        \hline
        Reasoning Segmentation & Image & ReasonSeg &
        \textit{You need to perform Reasoning Segmentation on the image according to the Text Prompt.} &
        \textit{"The person who appears to have already won in the battle"} \\
        \hline
        Referring Video Object Segmentation & Video & Ref-YouTube-VOS, etc. &
        \textit{You need to perform Referring Video Object Segmentation on the video according to the Text Prompt.} &
        \textit{"A duck is held by a person with her both hands"} \\
        \hline
        Reasoning Video Object Segmentation & Video & ReVOS &
        \textit{You need to perform Reasoning Video Object Segmentation on the video according to the Text Prompt.} &
        \textit{"Which person is in the leading position?"} \\
        
        \bottomrule
    \end{tabularx}
\end{table*}

\paragraph{FLOPs Computation.}
We compute FLOPs for each component using closed-form expressions based on transformer operations. Nonlinear activations, biases, and normalization layers are omitted for simplicity. The estimates are as follows:

\begin{itemize}
  \item \textbf{Language model:} For $L$ layers, hidden size $d$, and intermediate FFN dimension $d_\mathrm{int}$:
  \begin{align}
    \mathrm{FLOPs}_\mathrm{attn} &= 3Sd^2 + 2S^2d + Sd^2, \notag\\
    \mathrm{FLOPs}_\mathrm{ffn}  &= 2Sd \cdot d_\mathrm{int}, \notag\\
    \mathrm{FLOPs}_\mathrm{LM}   &= L(\mathrm{attn} + \mathrm{ffn}) + Sd|\mathcal{V}|. \notag
  \end{align}

  \item \textbf{Vision encoder:} Given $N$ patches, hidden size $d_v$, and $L_v$ layers:
  \[
  \mathrm{FLOPs}_\mathrm{vision} = \sum_{\ell=1}^{L_v} \left(6Nd_v^2 + 2N^2d_v\right) + Nd_v d.
  \]

  \item \textbf{Token pruning:} For input $V$ and pruned $V'$ tokens:
  \[
  \mathrm{FLOPs}_\mathrm{prune} = 2Vd + \frac{VV'd}{10} + V'd.
  \]

  \item \textbf{Mask decoder:} Using $Q$ object queries, hidden dim $d_m$, and $L_d$ layers:
  \[
  \begin{aligned}
  \mathrm{FLOPs}_\mathrm{mask} =\; & 
  L_d \cdot (12Qd_m^2 + 2Q^2d_m + 2QV'd_m) \\
  & + Qd_m V'.
  \end{aligned}
  \]

  \item \textbf{Temporal reasoning:} With $Q_t$ temporal queries, $F$ frames, and $L_t$ layers:
  \[
  \mathrm{FLOPs}_\mathrm{temporal} = L_t (Q_t F d^2 + 4Q_t d^2).
  \]

  \item \textbf{VMTF fusion:} For fusion dim $d_f$ and $L_f$ layers:
  \[
  \mathrm{FLOPs}_\mathrm{vmtf} = L_f (T_\mathrm{eff} d d_f + V'd_v d_f + 2T_\mathrm{eff}V'd_f).
  \]
\end{itemize}

Each value corresponds to a single forward pass. For every configuration, we compute FLOPs under both unpruned ($V' = V$) and pruned settings, and report the reduction percentage.

\paragraph{Model Hyperparameters.}
All FLOPs estimates are based on the default InstructSeg configuration:

\begin{itemize}
  \item \textbf{Language model:} $d = 2560$, $d_\mathrm{int} = 10240$, $L = 32$ layers, vocabulary size $|\mathcal{V}| = 51200$.
  \item \textbf{Vision encoder:} $d_v = 1152$, $N = 729$ patches from $384 \times 384$ input, $L_v = 27$ layers.
  \item \textbf{Mask decoder:} $Q = 100$ queries, $d_m = 256$, $L_d = 9$ layers.
  \item \textbf{Temporal module:} $Q_t = 128$ queries, $L_t = 3$ layers, $F = 4$ reference frames.
  \item \textbf{VMTF fusion:} $d_f = 1024$, $L_f = 3$ layers.
  \item \textbf{Token pruning:} Hidden dim $d = 2560$.
  \item \textbf{Fixed tokens:} $T_\mathrm{fixed} = 100$.
\end{itemize}

These settings are used consistently across all datasets without per-task tuning.

\subsection{Instruction Templates and Dataset Construction}
\label{subsec:data_instruction}

Following \textit{InstructSeg}~\cite{wei2024instructseg}, we use task-specific instruction templates to construct language prompts. Each segmentation task is paired with a distinct instruction format based on its modality and reasoning type. These templates remain consistent across training and evaluation. Table~\ref{tab:prompt_design} summarizes the instruction settings and representative examples.

\subsection*{Baseline Pruning Methods}

We compare with several state-of-the-art token pruning methods designed for general-purpose vision-language models. Specifically, we include ToM~\cite{bolya2022token}, VisionZip~\cite{yang2025visionzip}, VisPruner~\cite{zhang2024beyond}, and DivPrune~\cite{divprune}, all of which are compatible with frozen architectures and do not require retraining.

To ensure fair comparison, we unify the application point for all methods: visual token pruning is applied at the output of the SigLIP encoder. This allows consistent input representation into the language model. For attention-based methods (e.g., VisionZip), we extract attention weights directly from SigLIP. This differs slightly from the original setup, which typically relies on CLIP, but preserves each algorithm core mechanism. Since these pruning method were originally designed for Visual understanding tasks such as VQA and used primarily in LLaVA or LLama architecture MLLM, minor adaptation is necessary to evaluate them in the IVS based on MLLMs setting. We exclude other pruning baseline methods for several reasons:
\begin{itemize}
  \item \textbf{Retraining-dependent methods}, such as TokenPacker~\cite{li2025tokenpacker} and M3~\cite{m3}, require end-to-end optimization and introduce structural changes, which are incompatible with frozen models.
  \item \textbf{Temporal-incompatible designs}, such as FastV~\cite{fastv}, assume architectures tailored to video encoders, which differ significantly from our setup.
\end{itemize}

In summary, we focus on general, transferable, one-shot pruning methods that can be fairly applied to the unified visual token stream without requiring architecture-specific or fine-tuning.

\section{Visualization and Experiment Result Analysis}

\subsection{Image-Based IVS Pruning Analysis}

\begin{figure*}[t]
  \centering
  \includegraphics[width=0.9\textwidth]{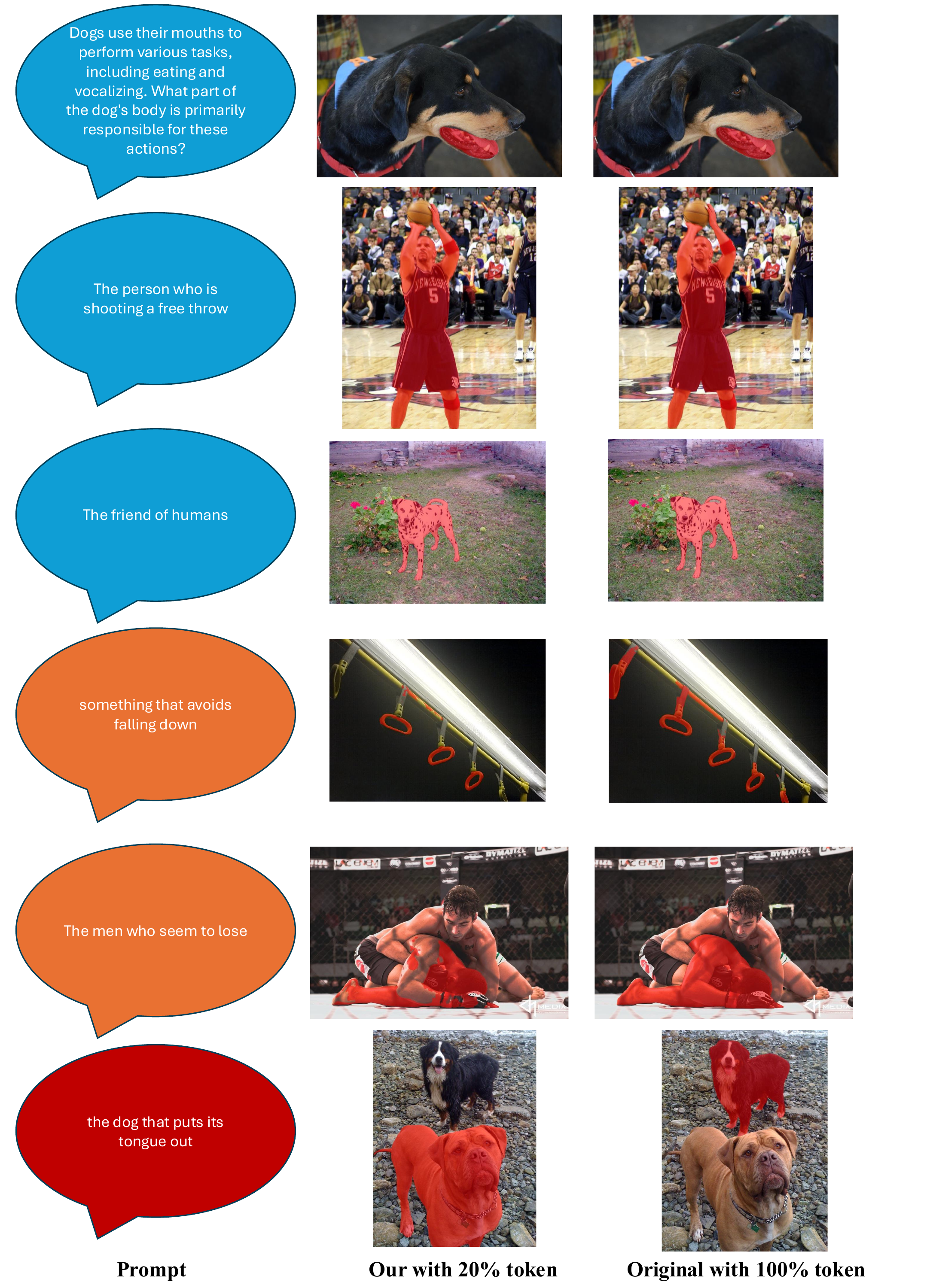}
  \caption{
    Qualitative results under 20\% visual token pruning and comparing with using original 100\% visual token. Blue prompt boxes indicate successful cases where segmentation closely matches the full-token model. Orange-yellow boxes denote partial successes where overall object segmentation is preserved but fine-grained details are lost or imprecise. Red boxes represent failure cases, often involving multiple objects or complex spatial layouts, where the model fails to accurately localize or identify the target object.
    }
  \label{fig:token_pruning_results}
\end{figure*}

To better understand how token pruning affects segmentation quality in practice, we examine qualitative results using only 20\% of the visual tokens. As shown in Figure~\ref{fig:token_pruning_results}, we compare outputs from our pruned model with those generated using the full token set across a range of visual inputs and prompts.
In the top examples with blue prompt bubbles, the model performs well despite heavy pruning. For instance, in the basketball image, the player is clearly segmented with accurate boundaries. Similarly, in both dog-related prompts, the model correctly identifies the dog with limited information. These results suggest that when objects are visually prominent and well separated from the background, a small number of well-distributed tokens is often enough to support good segmentation.
In the middle group, marked with orange bubbles, the segmentation is generally correct but misses some details. For example, in the subway handle image, several handles are not segmented. In the wrestling example, parts of the bodies are either incomplete or blurred. These cases indicate that while the model still captures the main object, it struggles with fine structures when fewer tokens are available.
The last row, highlighted in red, shows a failure case. The prompt asks for a “dog with its mouth open,” but the model only segments the larger black dog, completely missing the smaller beige dog that more closely matches the prompt. This likely results from insufficient token coverage in that region, suggesting that static selection may not adapt well to scenes with multiple similar objects or complex layouts.

Future pruning method could benefit from combining spatial coverage with semantic information, such as attention or uncertainty signals, to better identify which regions matter. Adapting the token budget based on input complexity or using multi-scale sampling might also help retain more useful information. Overall, while static pruning works well in many settings, more adaptive methods may be needed for robust performance across diverse visual inputs.

\subsection{Video-Based IVS Pruning Analysis}
\begin{figure*}[t]
  \centering
  \includegraphics[width=1.0\textwidth]{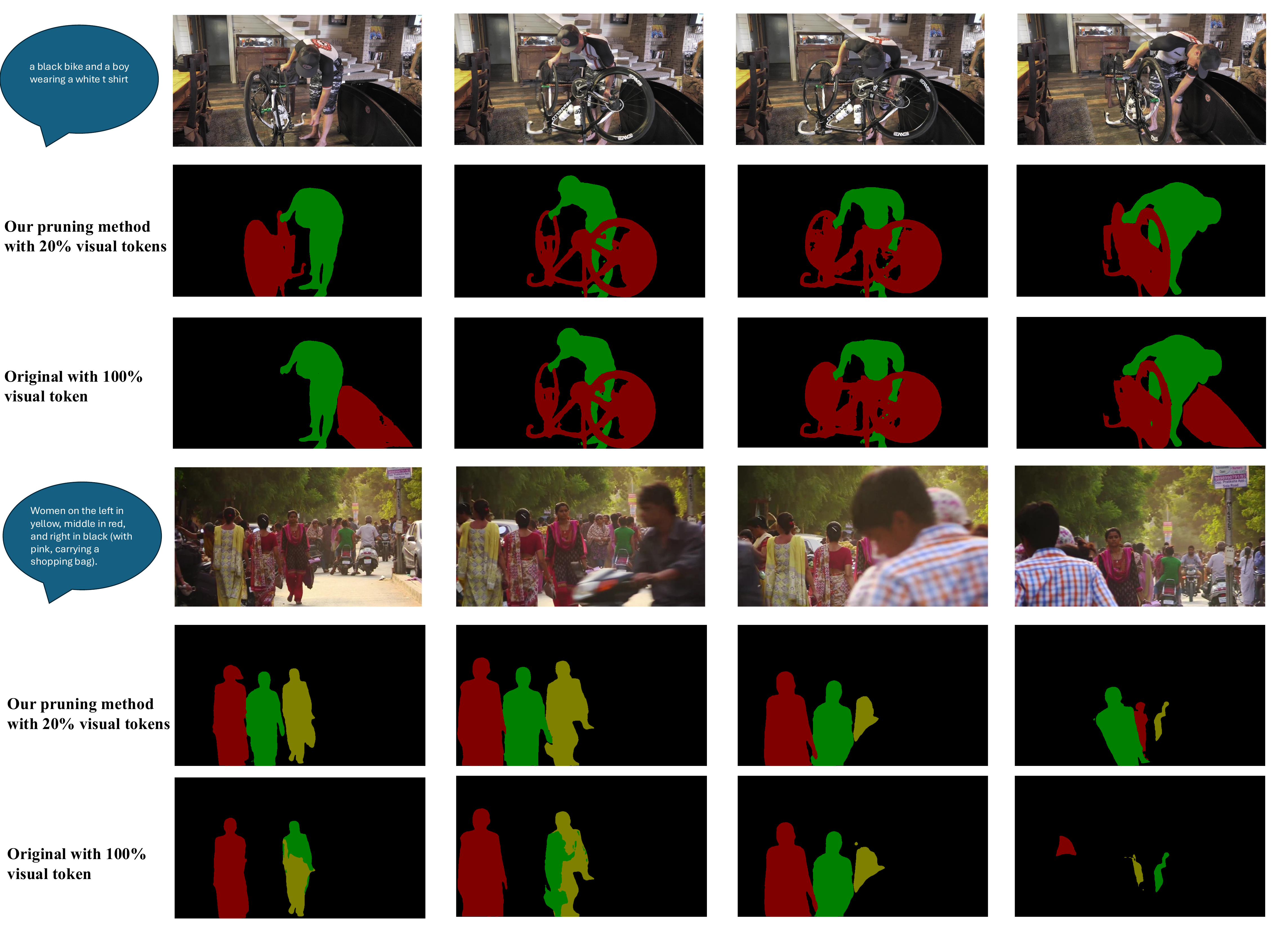}
  \caption{
    Video segmentation results using 20\% visual token pruning. Each example compares our pruned model (top row in each pair) with the original model using 100\% tokens (bottom row). In the first sequence, showing a person and a bicycle, pruning reduces clutter and leads to better segmentation in some frames. In the second sequence, which includes multiple people in a crowd, the pruned model produces better spatial separation between foreground and background.
    }
  \label{fig:video_pruning_results}
\end{figure*}

\begin{figure*}[t]
  \centering
  \includegraphics[width=1.0\textwidth]{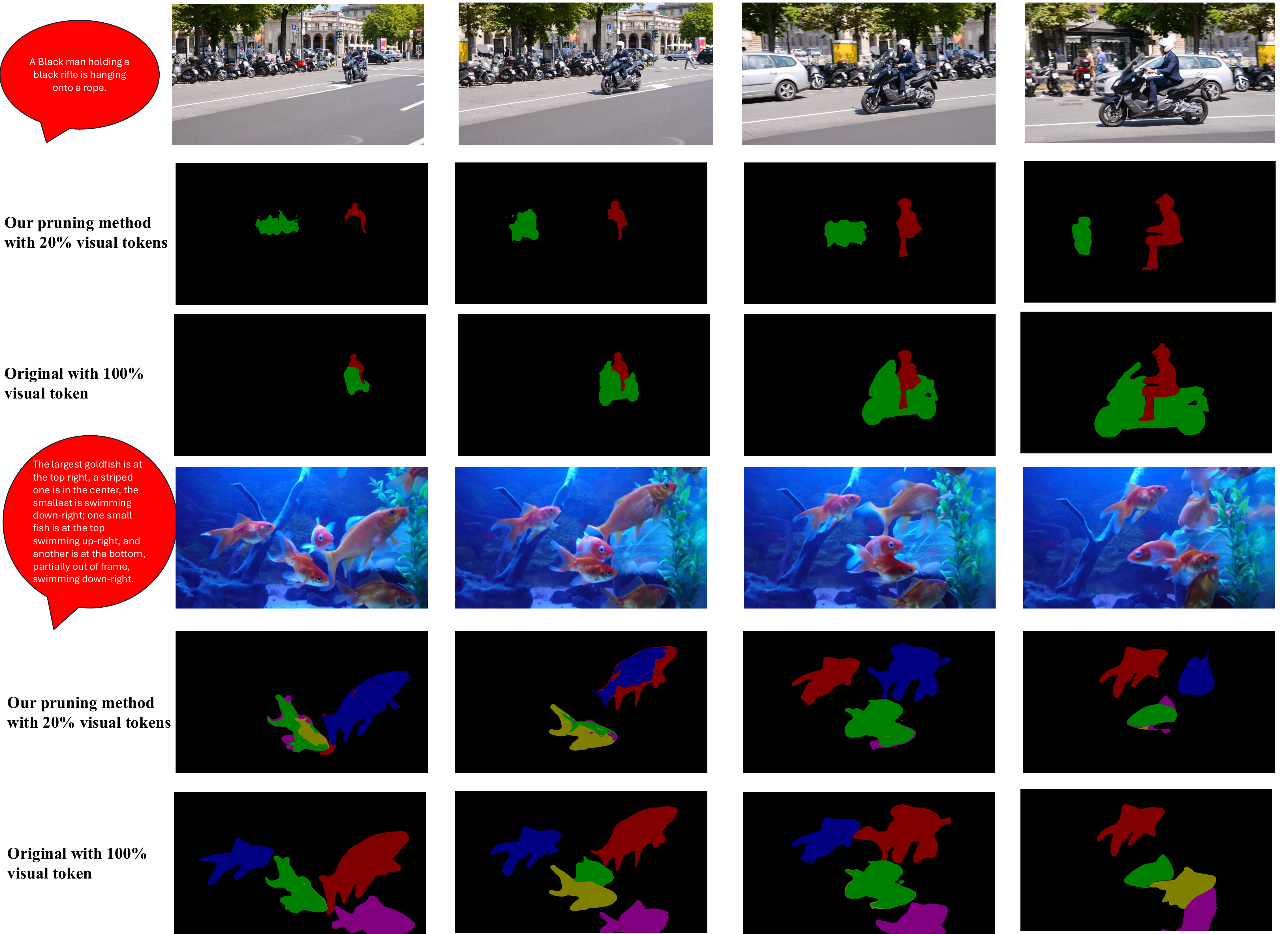}
  \caption{
    Failure cases of our visual token pruning method under 20\% token budget. In the top sequence, involving fast-moving motorcycles, the model fails to consistently segment both the rider and the vehicle, likely due to motion blur and temporal misalignment. In the bottom sequence, the scene contains multiple fish with highly similar appearances. The pruned model struggles to preserve instance separation and fails to segment all targets accurately. These examples illustrate that pruning method can underperform in cases of rapid motion or when many visually similar objects appear in the same frame.
    }
  \label{fig:failure_cases}
\end{figure*}

From previous experiment(Refer to table 2 and 3 in main manuscript), we find that pruning visual tokens to 20\% introduces relatively minor performance degradation in video-based IVS comparing to image-based IVS. This observation prompts a closer examination of the underlying factors contributing to this robustness. To this end, we analyze qualitative results in Figure~\ref{fig:video_pruning_results}, comparing segmentation outputs from models using full and pruned token sets across representative video frames.
Our findings suggest that the pruned model preserves segmentation quality despite the significant reduction in token count. In certain instances, pruning even improves boundary sharpness by suppressing irrelevant or redundant visual information and also shown better segmentation performance. This effect is particularly noticeable in scenes characterized by static backgrounds or repetitive textures, where token reduction helps eliminate noise that may otherwise interfere with accurate segmentation.
A key factor contributing to this resilience is the high degree of temporal redundancy inherent in video data. Consecutive frames often share similar visual content, enabling the model to leverage temporal continuity to maintain spatial coherence and semantic consistency across frames. As a result, the model can tolerate missing or reduced spatial cues in individual frames by drawing upon context from neighboring frames. This temporal smoothing effect mitigates the negative impact of token pruning and makes video a particularly suitable modality for token pruning-efficient modeling.
Nevertheless, the limitations of the current approach become evident in more challenging scenarios. As shown in Figure~\ref{fig:failure_cases}, the method struggles with rapid motion and fine-grained object discrimination. In the first example, fast-moving objects such as motorcycles exhibit motion blur and large frame-to-frame variation, which disrupt the model’s ability to consistently retain relevant tokens. In the second example, dense scenes involving multiple, visually similar objects (e.g., fish) pose difficulty for token selection strategies that prioritize spatial coverage over semantic distinctiveness.

These observations suggest that future pruning strategies should move beyond static heuristics and incorporate temporal dynamics, motion sensitivity, and object-level saliency into the selection process. Learning-based token prioritization mechanisms, possibly informed by task objectives or attention distributions, may offer a more adaptive and effective approach for maintaining segmentation performance under constrained token budgets.

\subsection{Dataset Comparison: DAVIS vs. YouTube-VOS}

Our main experiments (main manuscript) reveal that pruning leads to a larger performance drop on Refer-DAVIS17 compared to Refer-YouTube-VOS, even under the same token retention ratio. To better understand this discrepancy, we summarize key dataset statistics in Table~\ref{tab:dataset_stats}, focusing on factors that may influence pruning robustness.
\begin{table}[t]
\centering
\caption{Comparison of Refer-DAVIS17 and Refer-YouTube-VOS datasets.}
\label{tab:dataset_stats}
\resizebox{0.95\linewidth}{!}{
\begin{tabular}{l@{\hskip 10pt}c@{\hskip 10pt}c}
\toprule
\textbf{Statistic} & \textbf{Refer-DAVIS17} & \textbf{Refer-YouTube-VOS} \\
\midrule
Number of Videos & 90 & 3975 \\
Number of Objects & 205 & 7451 \\
Number of Expressions & 1544 & 27899 \\
Avg. Objects per Video & 2.3 & 1.87 \\
Avg. Expressions per Object & 7.53 & 3.74 \\
Avg. Frames per Sequence & 67.4 & 126.0 \\
Annotation Density & Dense (every frame) & Sparse (5 fps) \\
Category Diversity & Low (e.g., people, dogs) & High (animals, vehicles, tools) \\
\bottomrule
\end{tabular}
}
\end{table}
Refer-DAVIS17 features shorter sequences with dense frame-level annotations and highly focused object categories. Each object is described by multiple referring expressions and segmented in every frame, making fine-grained localization critical. Consequently, pruning even a small number of important tokens can noticeably affect segmentation boundaries or object completeness.
In contrast, Refer-YouTube-VOS consists of longer, more diverse videos with sparser annotations and greater temporal redundancy. This structure provides the model with more contextual support across frames, allowing it to compensate for missing details caused by pruning. These properties make YouTube-VOS more resilient to token reduction.

Overall, the differences in temporal structure, annotation density, and content diversity help explain the dataset-specific behavior observed in our experiments. These insights also motivate future work on dataset-aware pruning strategies that can adapt to the demands of different video understanding tasks.

\clearpage
\end{document}